\documentclass[runningheads]{helpers/llncs}
\usepackage[T2A,T1]{fontenc}
\usepackage[utf8]{inputenc}
\usepackage[russian,english]{babel}
\usepackage{graphicx}
\usepackage{xcolor}
\usepackage{hyperref}
\usepackage{caption}
\usepackage{amsfonts}
\usepackage{amsmath}
\usepackage{amssymb}
\usepackage{booktabs}
\usepackage{tabularx}
\usepackage{multirow}

\newcommand{\fat}[1]{\ifmmode\bm{#1}\else\textbf{#1}\fi}

\newcommand{\set}[1]{\mathbb{#1}}

\newcommand{\tens}[1]{\mathcal{#1}}

\newcommand{\func}[1]{\textsf{#1}}

\newcommand{\tran}[1]{\func{T}[#1]}
\begin{document}
\title{
    Translate your gibberish: black-box adversarial attack on machine translation systems
}
\titlerunning{Black-box adversarial attack on machine translation systems}

\author{
%Anonymous authors
    Andrei Chertkov\inst{1, 2}\orcidID{0000-0001-9990-6598}
    \and
    Olga Tsymboi\inst{3, 4}\orcidID{0000-0002-8078-1876}
    \and
    Mikhail Pautov\inst{1}\orcidID{0000-0003-0438-6361}
    \and
    Ivan Oseledets\inst{1,2,5}\orcidID{0000-0003-2071-2163}
}
\authorrunning{A. Chertkov et al.}

\institute{
    Skolkovo Institute of Science and Technology, Moscow, Russia
    \email{\{a.chertkov,mikhail.pautov,i.oseledets\}@skoltech.ru}
    \and
    Institute of Numerical Mathematics, Russian Academy of Sciences
    \and
    Moscow Institute of Physics and Technology, Moscow, Russia
    \email{tsimboy.oa@phystech.edu}
    \and
    Sber AI Lab, Moscow, Russia
    \and
    AIRI, Moscow, Russia
}

\maketitle

\begin{abstract}
    Neural networks are deployed widely in natural language processing tasks on the industrial scale, and perhaps the most often they are used as compounds of automatic machine translation systems. In this work, we present a simple approach to fool state-of-the-art machine translation tools in the task of translation from  Russian to English and vice versa. Using a novel black-box gradient-free tensor-based optimizer, we show that many online translation tools, such as Google, DeepL, and Yandex, may both produce wrong or offensive translations for nonsensical adversarial input queries and refuse to translate seemingly benign input phrases. This vulnerability may interfere with understanding a new language and simply worsen the user's experience while using machine translation systems, and, hence, additional improvements of these tools are required to establish better translation.

\keywords{
    Natural language processing
    \and
    Machine translation
    \and
    Adversarial attack
    \and
    Black-box optimization
}
\end{abstract}
\section{Introduction}

Adversarial perturbations are carefully crafted modifications of the input that are imperceptible for humans but force a machine learning model to perform poorly.
Initially discovered in the domain of computer vision~\cite{DBLP:journals/corr/SzegedyZSBEGF13,DBLP:journals/corr/GoodfellowSS14}, where imperceptibility is attained by restricting the norm of additive perturbation, they were later extended to the natural language processing (NLP).
Since the nature of language is discrete, the imperceptibility in NLP is attained either on the character-level~\cite{DBLP:conf/coling/EbrahimiLD18,DBLP:conf/acl/EbrahimiRLD18}, where only few characters in a word are subject to change, or on the word-level~\cite{DBLP:conf/conll/BlohmJSYV18,DBLP:conf/aaai/ChengYCZH20}, where the words are allowed to be replaced only by the semantically similar words (e.g., by synonyms). 

However, machine translation (MT) systems are known to be vulnerable to adversarial examples with relaxed imperceptibility~\cite{chen2022should}.
More than that, apart from sensitivity to imperceptible adversarial examples, MT may both produce meaningful translations for nonsensical gibberish input queries and refuse to translate seemingly benign input phrases.
This unpredictable behavior may not only interfere with understanding a new language but also may lead to serious problems (e.g., several years ago Facebook’s MT system mistranslated an Arabic phrase meaning ``good morning'' as ``attack them'' which led to a wrongful arrest~\cite{berger2017israel,ebrahimi2018adversarial}). Hence, understanding the unpredictable behavior of these systems is an essential step for improving the robustness of machine translation and, as a result, for preventing such incidents. 

In this work, we investigate the stability and behavior of MT systems for inputs with low likelihood.
We consider three major well-known online translators DeepL
%\footnote{
%    \url{https://www.DeepL.com/translator}.
%},
Google,
%\footnote{
%    \url{https://translate.google.com}.
%}
and Yandex,
%\footnote{
%\url{https://translate.yandex.ru}.
%},
and set the task of automatically finding an input in Russian representing an arbitrary set of letters of a given length (not a word), which, however, leads to a meaningful translation into English (a word or set of words).
We formulate it as a problem of maximizing the difference between the perplexity~\cite{sadrizadeh2023transfool} of the translation and the source text, and we apply GPT-2~\cite{radford2019language} to define the perplexity of the input and output sequences.
For a search of the best combination of input symbols we use the new optimization method PROTES\footnote{
    We use the code from \url{https://github.com/anabatsh/PROTES}.
}~\cite{batsheva2023protes}, which is based on the low-rank tensor train (TT) decomposition~\cite{oseledets2011tensor} and can efficiently perform gradient-free multivariate discrete optimization.
For all three considered MT systems, we obtained a set of seven-letter inputs in Russian that are not words, which, however, lead to a translation representing a word or set of words in English.
Hereafter, for the sake of brevity, we will refer to such inputs as \emph{hallucinogens}.
What is an intriguing, both manual and automatic combinations of the obtained hallucinogens, as it turned out, allows getting a variety of valid English phrases.
Moreover, some of these phrases turn out to be examples of adversarial attacks (detected so far only for the DeepL translator).
When trying to translate them back into Russian, the translator produces significantly incorrect results (garbage word combinations or even a blank translation string).
To summarize, our contributions are the following:
\begin{itemize}
    \item
        We develop a new black-box optimization method for the automatic generation of low-likelihood input sequences (``hallucinogens'')  with high translation likelihood for MT systems based on the perplexity estimation of the input and output sequences. %and the PROTES optimization method in the TT-format.
    \item
        We demonstrate that it is possible to use this approach for black-box adversarial attacks on MT systems since the corresponding translation results for a set (phrase) of hallucinogens often correspond to the ``instability points'' of the system and lead to invalid backward translation.
    \item
        We apply\footnote{
            The program code and all results with the supporting screenshots are available in our public repository \url{https://github.com/AndreiChertkov/TranFighterPro}.
        } the proposed approach for major online translators DeepL, Google, and Yandex, find an extensive set of hallucinogens and their combinations for all three translators, and demonstrate the possibility of an adversarial attack on the DeepL system.
\end{itemize}
\section{Method}

\begin{figure}[t!]
\includegraphics[width=\textwidth]
    {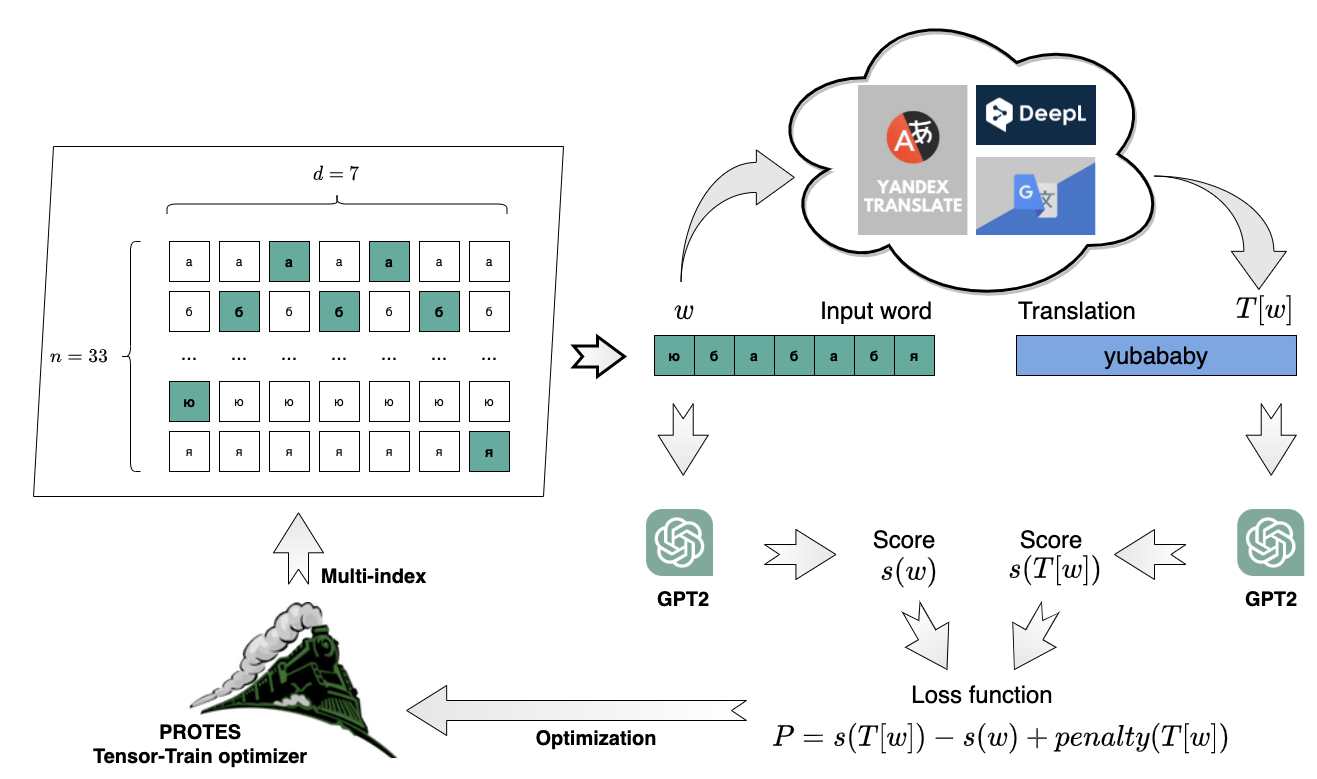}
\caption{
    Proposed approach for the search of the ``hallucinogens''.
}
\label{fig:method}
\end{figure}

Our approach is presented in Figure~\ref{fig:method} and is based on the idea of searching for $d$-letter combinations $w = (w_1, w_2, \ldots, w_d)$ in the source language that are the least similar to the existing words (gibberish or “hallucinogens”), however correctly translatable into the target language as  $\tran{w}$. 
Without loss of generality, we have chosen Russian as the source language (it has $n = 33$ letters of the alphabet), English language as the target language (it has $n_t = 26$ letters of the alphabet), and $d = 7$.

To assess the quality (score) of a word or phrase, we use perplexity~\cite{sadrizadeh2023transfool}:
\begin{equation}\label{eq:score}
\func{s}(w) =
    \func{exp}\left[
        -\frac{1}{d}
        \sum_{i=1}^{d}
            \log p_{\theta}(w_i | w_{<i})
    \right],
\end{equation}
where $p_{\theta}(w_i | w_{<i})$ is the log-likelihood of the i-th token conditioned on the preceding tokens according to the pre-trained GPT-2 model.
It can be thought of as an evaluation of the model’s ability to predict among the set of specified tokens in a corpus.
The value $\func{s}(w)$ is non-negative, for the most common words it is close to zero, and for the gibberish, it is expected to be a large positive number.

To maximize the difference between the perplexity of the translation $\tran{w}$ and the source text $w$ we introduce the following loss function:
\begin{equation}\label{eq:optimizer_loss_level1}
P(w) =
    \func{s}(\tran{w}) -
    \func{s}(w) +
    \func{penalty}(\tran{w}),
\end{equation}
where $\func{penalty}(\tran{w})$ is a penalty term, which is equal to a large positive number for the case when the translation is too short (less than $5$ characters) or contains stop characters (various non-letter characters); otherwise it is zero.

We search for minimum of~\eqref{eq:optimizer_loss_level1} in terms of the discrete optimization problem for an implicitly given $d$-dimensional array $\tens{P} \in \set{R}^{n \times n \times \ldots \times n}$:
\begin{equation}\label{eq:tensor}
\tens{P}[i_1, i_2, \ldots, i_d] =
    P(w),
\quad
w = (
    A[i_1], A[i_2], \ldots, A[i_d]
),
\end{equation}
where $[i_1, i_2, \ldots, i_d]$ is a multi-index, $A$ is the alphabet, and $A[i_k]$ if the $i_k$-th symbol of the alphabet.
For example, as shown in Figure~\ref{fig:method}, for the multi-index $[32,\,2,\,1,\,2,\,1,\,2,\,33]$ we get the word $w$ ``\foreignlanguage{russian}{юбабабя}'' in Russian.

To find the ``hallucinogen'' $\hat{w}$ which minimizes the loss function~\eqref{eq:optimizer_loss_level1}, we use the global optimization method PROTES.
It is based on the low-rank tensor train (TT) decomposition~\cite{oseledets2011tensor,cichocki2016tensor,cichocki2017tensor,sozykin2022ttopt,chertkov2022optimization}, which allows bypassing the curse of dimensionality problem\footnote{
    The complexity of algorithms in the TT-format (e.\,g., element-wise addition, multiplication, solution of linear systems, convolution, integration, etc.) turns out to be polynomial in dimension and mode size, and it makes TT-decomposition extremely popular in a wide range of applications, including computational mathematics and machine learning.
}.
% In the last few years, several new discrete optimization algorithms based on the TT-format have been proposed~\cite{sozykin2022ttopt,selvanayagam2022global,shetty2022tensor,chertkov2022optimization}.
% Due to the use of a low-rank format, they are particularly effective for highly multidimensional optimization problems.
% However, the most suitable in our case is the PROTES method, which 
The method operates with a multidimensional discrete probability distribution in the TT-format, followed by efficient sampling from it and updating its parameters by stochastic gradient ascent to approximate the minimum or maximum in a better way.
We save the request history of the optimization method and, at the end of its run, we form a set of hallucinogens $\hat{w}^{(1)}, \hat{w}^{(2)}, \ldots, \hat{w}^{(m)}$ ($m$ here is a number of requests for a translator, i.e. computational budget), ordered by the value of the loss function.

It is worth mentioning that the described method does not generate adversarial examples per se (i.e., it does not force mistranslation) but produces examples (hallucinogens) that are translatable when they should not be.
However, it turns out to be an interesting empirical fact that combinations of hallucinogens also lead to the emergence of translation artifacts, while, as we will show below, these artifacts can turn out to be long meaningful phrases in the target language.

Accordingly, in the second stage, we repeat the described optimization process, composing phrases of $d^{(2)}$ hallucinogens.
As the possible candidates, we select $n^{(2)}$ ($n^{(2)} \leq m$) top hallucinogens $\hat{w}^{(1)}, \hat{w}^{(2)}, \ldots, \hat{w}^{(n^{(2)})}$ from the result of the first stage.
Without loss of generality, we have chosen $d^{(2)} = 7$ and $n^{(2)} = 33$, i.e., the same values as in the first stage.
In this case, we use the loss function~\eqref{eq:optimizer_loss_level1} without the second term, i.e., we do not maximize the perplexity of the input text, since it is already composed of the hallucinogens.
Note that we can repeat this process an arbitrary number of times, getting longer and longer ``phrases'' from the hallucinogens.
\section{Experiments}

\begin{table}[]
\caption{
    Top-$33$ generated hallucinogens for DeepL translator.
}
\label{tbl:results_DeepL_level1}
    \centering
    \begin{tabular}{llr|llr|llr}
    \toprule
    Text & Translation & Loss & Text & Translation & Loss & Text & Translation & Loss \\
    \midrule
         \foreignlanguage{russian}{быелръъ} & formerly & -42.52 & \foreignlanguage{russian}{оощвишн} & Promotion & -26.86 & \foreignlanguage{russian}{гзйкщчж} & gzcjcj & -23.04\\
\foreignlanguage{russian}{пдлешйщ} & Synopsis: & -39.47 & \foreignlanguage{russian}{ощуъиъв} & Feelings & -25.08 & \foreignlanguage{russian}{ъоэсйьл} & Yoesyl & -22.33\\
\foreignlanguage{russian}{бысёъгч} & Quickly & -38.53 & \foreignlanguage{russian}{гбьъьиэ} & gbjie & -24.08 & \foreignlanguage{russian}{мжвлвфж} & mjvlvfj & -22.0\\
\foreignlanguage{russian}{чтьёиэе} & READ MORE & -37.2 & \foreignlanguage{russian}{рыьдяно} & snarky & -24.07 & \foreignlanguage{russian}{ктлтксь} & ktltx & -21.61\\
\foreignlanguage{russian}{щосющйе} & Synopsis: & -34.84 & \foreignlanguage{russian}{жьрэиэф} & zhreif & -23.64 & \foreignlanguage{russian}{фйвьжиы} & fyvji & -21.38\\
\foreignlanguage{russian}{быншийя} & former & -34.84 & \foreignlanguage{russian}{жцчыщцй} & Žučičky & -23.64 & \foreignlanguage{russian}{жаьйщсч} & zhayshch & -21.25\\
\foreignlanguage{russian}{зсзгвлэ} & ssgvle & -30.42 & \foreignlanguage{russian}{чёхёшьч} & What the fuck & -23.49 & \foreignlanguage{russian}{ккзёйьи} & kkzoyi & -20.78\\
\foreignlanguage{russian}{бгаьъэы} & bgaiy & -30.12 & \foreignlanguage{russian}{зжнмкьъ} & zznnmkj & -23.37 & \foreignlanguage{russian}{бфзскйт} & bfzskyt & -20.66\\
\foreignlanguage{russian}{дачэщйч} & Dachshund & -27.67 & \foreignlanguage{russian}{гмххъьн} & gmhxjn & -23.21 & \foreignlanguage{russian}{ыьбэъхс} & yybexx & -20.47\\
\foreignlanguage{russian}{бреощее} & Breaking & -27.5 & \foreignlanguage{russian}{жьрцэъо} & Jrceo & -23.19 & \foreignlanguage{russian}{ъйлбмфь} & ylbmfj & -20.27\\
\foreignlanguage{russian}{бжкльлш} & bjklsh & -27.21 & \foreignlanguage{russian}{бёацсжю} & boatsjue & -23.15 & \foreignlanguage{russian}{чъръпьм} & chirp & -20.23\\
    \bottomrule
    \end{tabular}
\end{table}

\begin{table}[]
\caption{
    Top-$33$ generated hallucinogens for Google translator.
}
\label{tbl:results_google_level1}
    \centering
    \begin{tabular}{llr|llr|llr}
    \toprule
    Text & Translation & Loss & Text & Translation & Loss & Text & Translation & Loss \\
    \midrule
    \foreignlanguage{russian}{ъувщжёь} & Knight & -50.18 & \foreignlanguage{russian}{штшнлхж} & Stitch & -35.53 & \foreignlanguage{russian}{ъокнёйф} & Continuity & -30.15\\
\foreignlanguage{russian}{бйввкшя} & Former & -48.27 & \foreignlanguage{russian}{гяшрьнп} & Gagarin & -33.98 & \foreignlanguage{russian}{ъфъыхлч} & Kommersant & -30.1\\
\foreignlanguage{russian}{дщижщяп} & Building & -45.13 & \foreignlanguage{russian}{здкънсп} & health & -33.39 & \foreignlanguage{russian}{птйдфдц} & PTDDC & -30.09\\
\foreignlanguage{russian}{мощыъпз} & Power & -43.64 & \foreignlanguage{russian}{ъыллщьн} & Kommersant & -32.24 & \foreignlanguage{russian}{йтдкцяе} & induction & -29.54\\
\foreignlanguage{russian}{ъыьгрвх} & Kommersant & -43.38 & \foreignlanguage{russian}{ътшлшэь} & Kommersant & -32.0 & \foreignlanguage{russian}{уясъцёь} & understanding & -29.29\\
\foreignlanguage{russian}{пёвюмыц} & first & -41.73 & \foreignlanguage{russian}{быошийя} & To be & -31.81 & \foreignlanguage{russian}{зсзгвлэ} & ZSZGLE & -29.28\\
\foreignlanguage{russian}{ъёефнся} & Currently & -41.19 & \foreignlanguage{russian}{доцшлны} & Associated & -31.69 & \foreignlanguage{russian}{ъфоъкцж} & Kommersant & -29.01\\
\foreignlanguage{russian}{ъжлхчлы} & Kommersant & -37.32 & \foreignlanguage{russian}{пщмёжны} & They are & -31.62 & \foreignlanguage{russian}{жхнаеыь} & grunts & -28.97\\
\foreignlanguage{russian}{ъоэсйьл} & Kommersant & -37.21 & \foreignlanguage{russian}{ъухвмгс} & Kommersant & -31.38 & \foreignlanguage{russian}{ъфкщтнэ} & Kommersant & -28.68\\
\foreignlanguage{russian}{вытёщдч} & priest & -37.05 & \foreignlanguage{russian}{ъбывзлц} & Kommersant & -30.8 & \foreignlanguage{russian}{ъныуазу} & Kommersant & -28.47\\
\foreignlanguage{russian}{бщагчёщ} & Passing & -36.29 & \foreignlanguage{russian}{бяёщжии} & beads & -30.24 & \foreignlanguage{russian}{гфоаььн} & fifajn & -28.38\\
    \bottomrule
    \end{tabular}
\end{table}

\begin{table}[]
\caption{
    Top-$33$ generated hallucinogens for Yandex translator.
}
\label{tbl:results_yandex_level1}
    \centering
    % \fontsize{9pt}{9pt}
% \selectfont
\setlength{\tabcolsep}{1pt}
    \begin{tabular}{llr|llr|llr}
    \toprule
    Text & Translation & Loss & Text & Translation & Loss & Text & Translation & Loss \\
    \midrule
    \foreignlanguage{russian}{здблоьп} & hello & -42.87 & \foreignlanguage{russian}{кмтсгфк} & kmtsgfc & -27.48 & \foreignlanguage{russian}{иьллтёу} & illteu & -24.03\\
\foreignlanguage{russian}{ьвднэйу} & Today & -42.15 & \foreignlanguage{russian}{иощсцйм} & ioschcym & -27.08 & \foreignlanguage{russian}{щаафечу} & right now & -23.68\\
\foreignlanguage{russian}{онуьлйц} & online & -40.44 & \foreignlanguage{russian}{нзеъёаь} & nzeea & -26.32 & \foreignlanguage{russian}{ъяляужь} & for the service & -23.41\\
\foreignlanguage{russian}{смэёыюш} & see also & -35.26 & \foreignlanguage{russian}{бмъчкьь} & bmchk & -26.1 & \foreignlanguage{russian}{нмьрщшт} & nmrsht & -23.33\\
\foreignlanguage{russian}{иысвщёы} & and more & -34.94 & \foreignlanguage{russian}{ъоэсйьм} & yoesm & -25.67 & \foreignlanguage{russian}{оэеыъьё} & oeeye & -23.16\\
\foreignlanguage{russian}{схисеъм} & scheme & -32.76 & \foreignlanguage{russian}{ъыклщьн} & kommersant & -25.56 & \foreignlanguage{russian}{йьаёьеб} & yaeeb & -23.1\\
\foreignlanguage{russian}{мощыъпз} & The~power~of~the & -31.2 & \foreignlanguage{russian}{бьвтюья} & byuya & -25.49 & \foreignlanguage{russian}{флжсйид} & fljsyid & -22.72\\
\foreignlanguage{russian}{кццжйхк} & kccjhk & -30.76 & \foreignlanguage{russian}{иьеьрёъ} & iyere & -25.48 & \foreignlanguage{russian}{пёыэулм} & peeulm & -22.67\\
\foreignlanguage{russian}{ътшмщэь} & kommersant & -30.54 & \foreignlanguage{russian}{ущйинъу} & pinyin & -25.22 & \foreignlanguage{russian}{бдлпроь} & bdlpro & -22.59\\
\foreignlanguage{russian}{ъубщжёь} & kommersant & -27.58 & \foreignlanguage{russian}{шэьдкйя} & shadkya & -24.49 & \foreignlanguage{russian}{доцшлмь} & assoc . & -22.56\\
\foreignlanguage{russian}{ъььгрвх} & ygrvh & -27.56 & \foreignlanguage{russian}{ощуъиъв} & feeling & -24.03 & \foreignlanguage{russian}{ъныуазу} & kommersant & -22.53\\
    \bottomrule
    \end{tabular}
\end{table}

\begin{table}[]
\centering
\caption{
    Some examples for generated combinations of the hallucinogens for DeepL translator.
}
\label{tbl:results_DeepL_level2}
    \begin{tabular}{p{5.5cm}|p{6.5cm}}
    \toprule
    Text & Translation \\
    \midrule
%\foreignlanguage{russian}{ъоэсйьл гзйкщчж гбьъьиэ жьрэиэф зжнмкьъ бысёъгч бреощее}
%&
%The following is an example of the following
%\\
%\midrule
%\foreignlanguage{russian}{пдлешйщ фйвьжиы ккзёйьи гзйкщчж быншийя чтьёиэе ккзёйьи}
%&
%The first part of the story is the following
%\\
%\midrule
\foreignlanguage{russian}{жьрцэъо жьрцэъо ощуъиъв ъйлбмфь чтьёиэе ъйлбмфь зжнмкьъ}
&
Greetings from the Greetings Department of the Ministry of Foreign Affairs
\\
\midrule
\foreignlanguage{russian}{быншийя бгаьъэы ъоэсйьл чёхёшьч мжвлвфж рыьдяно гзйкщчж}
&
The formerly bogeyman is the one who is the most important person in the world.
\\
\midrule
\foreignlanguage{russian}{бреощее бысёъгч жаьйщсч жьрэиэф зсзгвлэ пдлешйщ оощвишн}
&
The main reason for this is that we have a lot of time and effort to get to the bottom of this \\
    \bottomrule
    \end{tabular}

\vspace{-0.3cm}
\end{table}

\begin{table}[]
    \centering
    \caption{
    Some examples for generated combinations of the hallucinogens for Google translator.
}
    \label{tbl:results_google_level2}
    \begin{tabular}{p{5.5cm}|p{6.5cm}}
    \toprule
    Text & Translation \\
    \midrule
\foreignlanguage{russian}{уясъцёь ъыллщьн пщмёжны ъныуазу йтдкцяе бщагчёщ ъёефнся}
&
understanding of the bang
\\
\midrule
\foreignlanguage{russian}{быошийя ъёефнся ъбывзлц ъжлхчлы быошийя йтдкцяе пёвюмыц}
&
I would have been the bungles of Kommersant Kommersant
\\
\midrule
%\foreignlanguage{russian}{вытёщдч доцшлны гфоаььн пёвюмыц бяёщжии ътшлшэь ъфоъкцж}
%&
%Stepschdch is an ahead of the National Academy of Education
%\\
%\midrule
%\foreignlanguage{russian}{уясъцёь дщижщяп штшнлхж пщмёжны пёвюмыц ъныуазу ъокнёйф}
%&
%understanding of the throat of the stamp
%\\
%\midrule
\foreignlanguage{russian}{вытёщдч доцшлны ъувщжёь бйввкшя пщмёжны ъыллщьн бяёщжии}
&
The priests of the Associate Professor Kommersant
\\
    \bottomrule
    \end{tabular}
\vspace{-0.3cm}
\end{table}
\begin{table}[t!]
    \centering
    \caption{
    Some examples for generated combinations of the hallucinogens for Yandex translator.
}
    \label{tbl:results_yandex_level2}
    \begin{tabular}{p{5.5cm}|p{6.5cm}}
    \toprule
    Text & Translation \\
    \midrule
\foreignlanguage{russian}{мощыъпз щаафечу йьаёьеб ощуъиъв нзеъёаь ощуъиъв иысвщёы}
&
The power of the heart is now being felt by the heart of the heart .
\\
\midrule
%\foreignlanguage{russian}{иысвщёы ъяляужь ощуъиъв мощыъпз ъныуазу ущйинъу ъныуазу}
%&
%and you will feel the power of the power of the power of the power /* and repeated ``of the power'' many times */
%\\
%\midrule
\foreignlanguage{russian}{ъяляужь иысвщёы иьллтёу оэеыъьё щаафечу мощыъпз ощуъиъв}
&
I will be able to feel the power of the heart.
\\
\midrule
%\foreignlanguage{russian}{бьвтюья иысвщёы бьвтюья здблоьп щаафечу шэьдкйя доцшлмь}
%&
%I'm sorry, I 'm sorry , I 'm sorry , I 'm sorry , /* and repeated ``I 'm sorry'' many times */
%\\
%\midrule
\foreignlanguage{russian}{ощуъиъв доцшлмь ъныуазу онуьлйц ьвднэйу здблоьп ьвднэйу}
&
I feel like I 'm on the right side of the right side of the right side of the right side of the right side of the right side of the right side
\\
    \bottomrule
    \end{tabular}
\vspace{-0.3cm}
\end{table}

\begin{figure}[t!]
\centering
\includegraphics[width=0.8\textwidth]
    {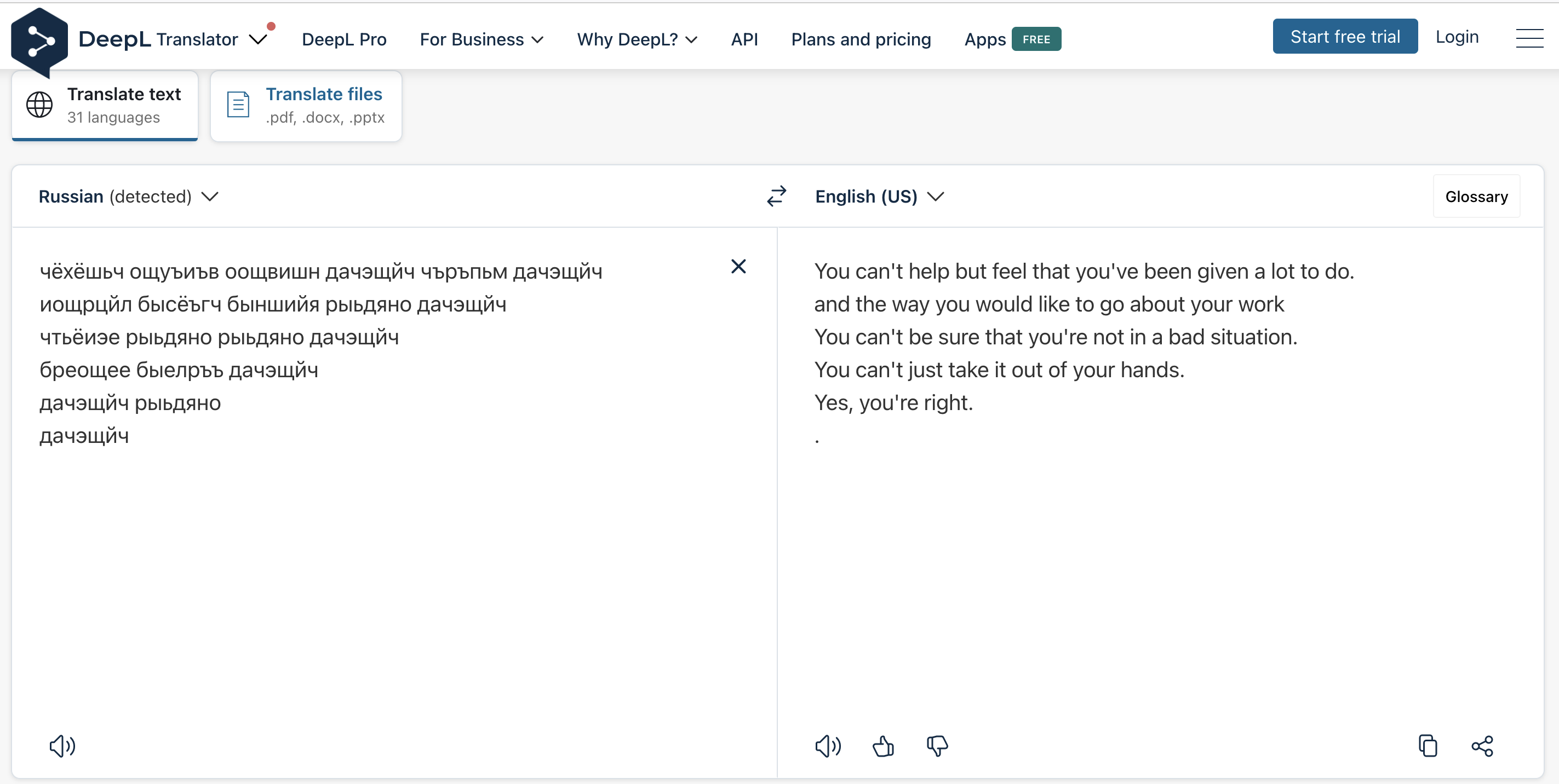}
\caption{
    Composition of hallucinogens for DeepL translator.
    %Russian text is ``\foreignlanguage{russian}{чёхёшьч ощуъиъв оощвишн дачэщйч чъръпьм дачэщйч
%иощрцйл бысёъгч быншийя рыьдяно дачэщйч
%чтьёиэе рыьдяно рыьдяно дачэщйч
%бреощее быелръъ дачэщйч
%дачэщйч рыьдяно
%дачэщйч}''.
}
\label{fig:demo_manual_DeepL}
\end{figure}

\begin{figure}[t!]
\centering
\includegraphics[width=0.8\textwidth]
    {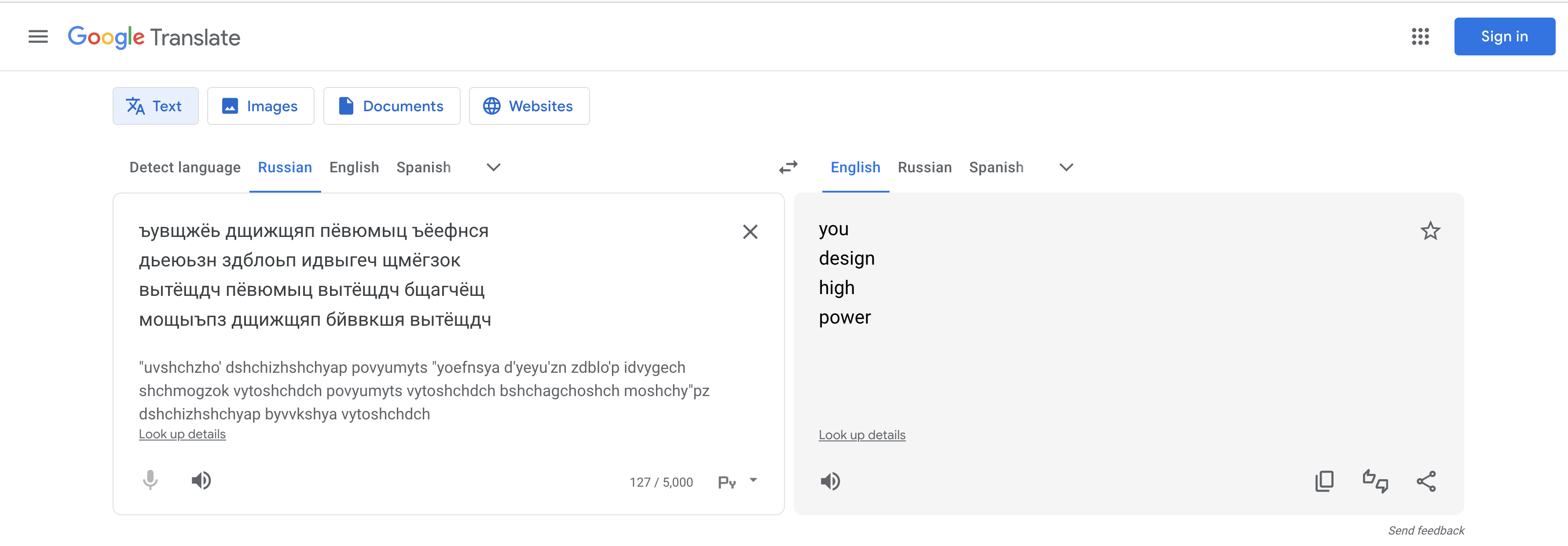}
\caption{
    Composition of hallucinogens for Google translator.
%    Russian text is ``\foreignlanguage{russian}{ъувщжёь дщижщяп пёвюмыц ъёефнся
%дьеюьзн здблоьп идвыгеч щмёгзок
%вытёщдч пёвюмыц вытёщдч бщагчёщ
%мощыъпз дщижщяп бйввкшя вытёщдч}''.
}
\label{fig:demo_manual_google}
\end{figure}

\begin{figure}[t!]
\centering
\includegraphics[width=0.8\textwidth]
    {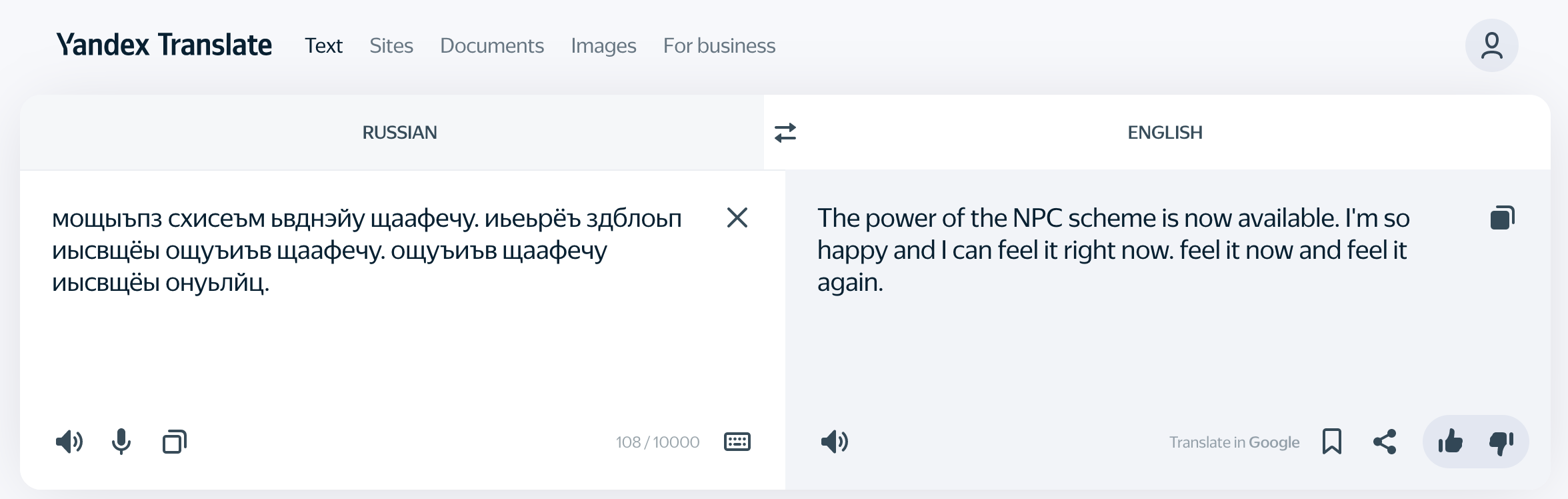}
\caption{
    Composition of hallucinogens for Yandex translator.
    %Russian text is ``\foreignlanguage{russian}{мощыъпз схисеъм ьвднэйу щаафечу. иьеьрёъ здблоьп иысвщёы ощуъиъв щаафечу. ощуъиъв щаафечу иысвщёы онуьлйц.}''.
}
\label{fig:demo_manual_yandex}
\end{figure}

\begin{figure}[t!]
\begin{center}
    \includegraphics[width=\textwidth]{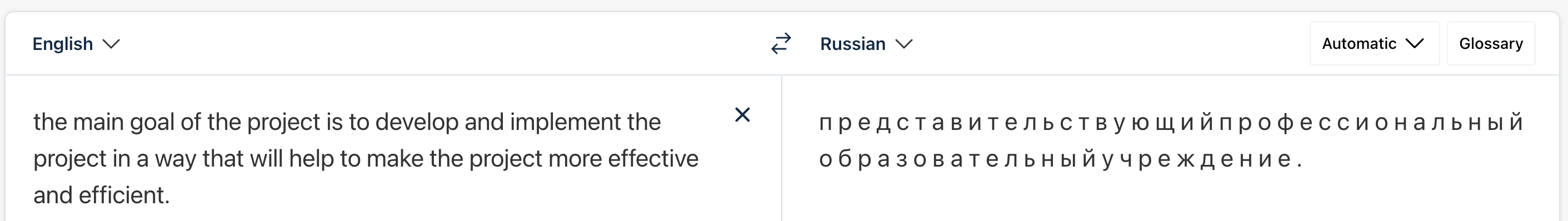}
\end{center}
\caption{
    Backtranslation results for the attack text 
    ``\foreignlanguage{russian}{фйвьжиы фйвьжиы пдлешйщ ккзёйьи гбьъьиэ жцчыщцй ктлтксь ыьбэъхс ъоэсйьл жьрцэъо мжвлвфж гзйкщчж жцчыщцй щосющйе ккзёйьи ккзёйьи фйвьжиы быншийя дачэщйч бысёъгч бёацсжю бысёъгч жцчыщцй жьрэиэф гмххъьн бёацсжю бгаьъэы чёхёшьч оощвишн бжкльлш бжкльлш щосющйе бгаьъэы дачэщйч ъоэсйьл пдлешйщ жцчыщцй жаьйщсч ъоэсйьл чёхёшьч бреощее ъйлбмфь бреощее бгаьъэы бжкльлш жьрэиэф ктлтксь ктлтксь бгаьъэы}''.
    % The text
    % ``the main goal of the project is to develop and implement the project in a way that will help to make the project more effective and efficient.''
    % is translated into Russian by DeepL translator incorrectly.
    The resulting Russian translation has the following meaning in English:
    ``representative professional educational institution''.
    \vspace{0.1cm}
}
\label{fig:DeepL_attack1}
\end{figure}

\begin{figure}[t!]
\includegraphics[width=\textwidth]
    {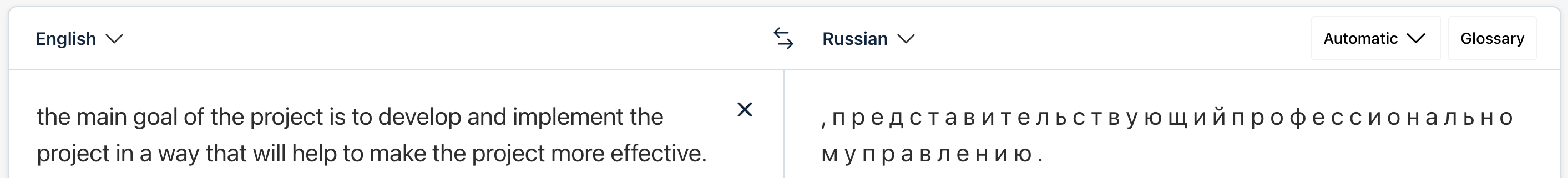}
\caption{
    % The text
    Backtranslation results for the attack text 
    ``\foreignlanguage{russian}{бёацсжю бгаьъэы гзйкщчж фйвьжиы дачэщйч бысёъгч ккзёйьи ъоэсйьл гзйкщчж гбьъьиэ жьрэиэф зжнмкьъ бысёъгч бреощее жьрцэъо быелръъ жаьйщсч бреощее зжнмкьъ чъръпьм ъйлбмфь ккзёйьи гзйкщчж гбьъьиэ зсзгвлэ жьрцэъо гзйкщчж чтьёиэе бысёъгч жцчыщцй жьрэиэф гмххъьн бёацсжю бгаьъэы чёхёшьч чёхёшьч ктлтксь бысёъгч ъоэсйьл быелръъ чёхёшьч гмххъьн жьрэиэф бжкльлш зсзгвлэ жьрцэъо бысёъгч бысёъгч бжкльлш}''.
    % ``the main goal of the project is to develop and implement the project in a way that will help to make the project more effective.''
    % is translated into Russian by DeepL translator incorrectly.
    The resulting Russian translation has the following meaning in English:
    ``representing professional management''.
    \vspace{0.1cm}
}
\label{fig:DeepL_attack2}
\end{figure}

\begin{figure}[t!]
\includegraphics[width=\textwidth]
    {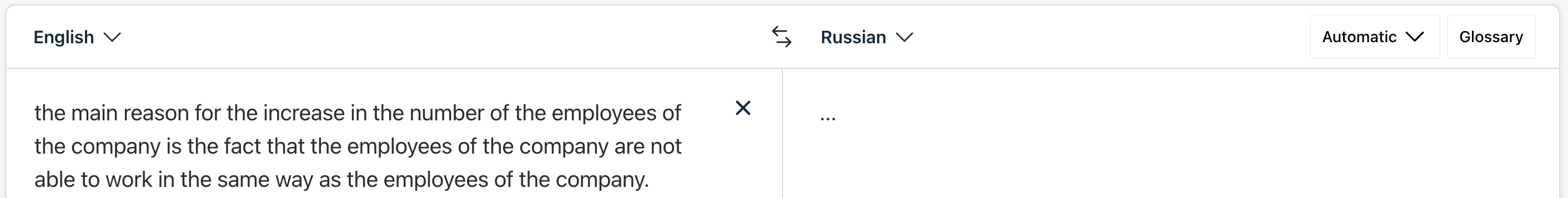}
\caption{
    % The text
    Backtranslation results for the attack text 
    % ``the main reason for the increase in the number of the employees of the company is the fact that the employees of the company are not able to work in the same way as the employees of the company.''
    % is translated into Russian by DeepL translator incorrectly.
    ``\foreignlanguage{russian}{рыьдяно рыьдяно фйвьжиы рыьдяно жьрэиэф щосющйе рыьдяно жцчыщцй фйвьжиы гбьъьиэ зсзгвлэ бгаьъэы рыьдяно ккзёйьи ктлтксь бфзскйт щосющйе пдлешйщ мжвлвфж рыьдяно гзйкщчж зсзгвлэ гзйкщчж гзйкщчж гбьъьиэ оощвишн гзйкщчж чёхёшьч пдлешйщ жцчыщцй жаьйщсч ъоэсйьл чёхёшьч бреощее ъйлбмфь ктлтксь бфзскйт щосющйе пдлешйщ мжвлвфж рыьдяно гзйкщчж чъръпьм чъръпьм ъйлбмфь пдлешйщ быншийя ощуъиъв ыьбэъхс}''.
    The resulting Russian translation is empty.
    \vspace{0.1cm}
}
\label{fig:DeepL_attack3}
\end{figure}

\begin{figure}[t!]
\includegraphics[width=\textwidth]
    {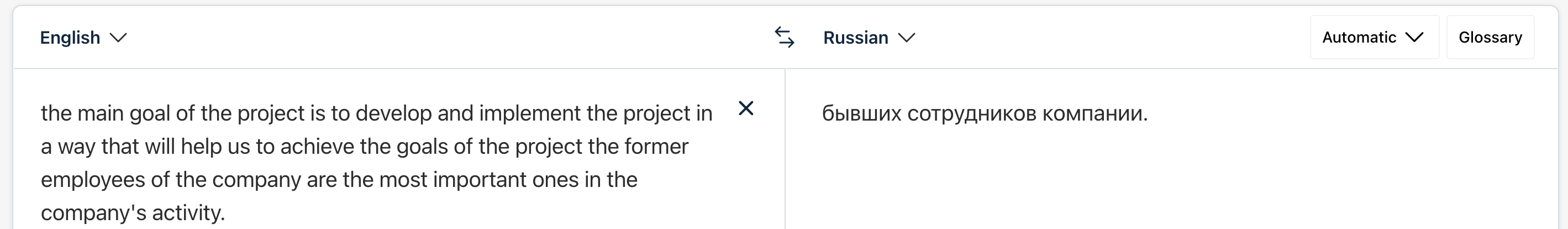}
\caption{
    % The text
    Backtranslation results for the attack text 
    % ``the main goal of the project is to develop and implement the project in a way that will help us to achieve the goals of the project the former employees of the company are the most important ones in the company's activity.''
    % is translated into Russian by DeepL translator incorrectly.
    ``\foreignlanguage{russian}{бёацсжю бгаьъэы гзйкщчж фйвьжиы дачэщйч бысёъгч ккзёйьи чёхёшьч ктлтксь бысёъгч ъоэсйьл быелръъ чёхёшьч гмххъьн ъоэсйьл ккзёйьи бжкльлш пдлешйщ рыьдяно жьрцэъо пдлешйщ бёацсжю зсзгвлэ бёацсжю чтьёиэе быншийя бжкльлш гзйкщчж чъръпьм чъръпьм ъйлбмфь пдлешйщ быншийя ощуъиъв ыьбэъхс бёацсжю бгаьъэы бреощее зжнмкьъ жаьйщсч ктлтксь ккзёйьи оощвишн бжкльлш бжкльлш щосющйе бгаьъэы дачэщйч ъоэсйьл}''.
    The resulting Russian translation has the following meaning in English:
    ``former employees of the company.''.
}
\label{fig:DeepL_attack5}
\end{figure}

\begin{figure}[t!]
\centering
\includegraphics[width=0.99\textwidth]
    {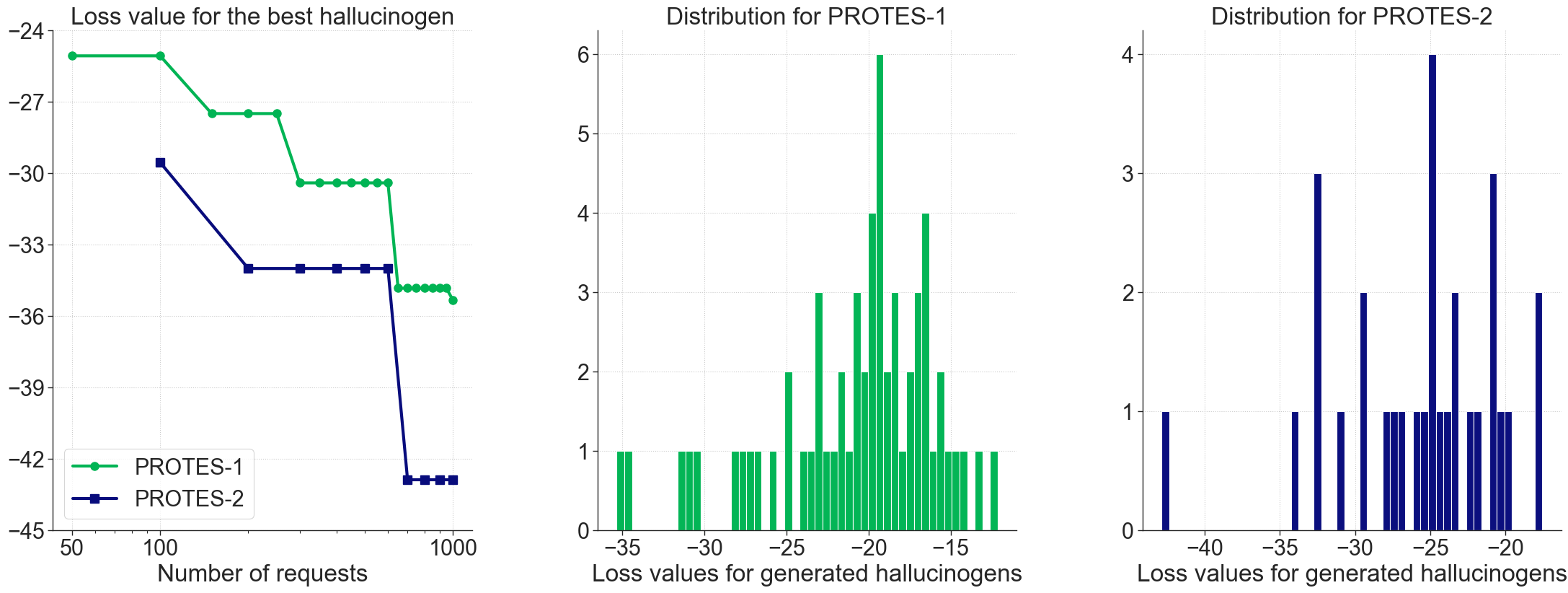}
\caption{
    The dependence of the found optimum on the number of requests to the DeepL online translator (on the graph on the left) and the distribution of results (on the graph in the center and on the right) for two optimizer configurations.
}
\label{fig:check}
\end{figure}

\begin{table}[t!]
\caption{
    The best generated hallucinogens for DeepL translator for each requested batch.
    The results for the two optimizer configurations with the batch size $50$ (PROTES 1) and $100$ (PROTES 2) are reported.
}
\label{tbl:results_deepl_check}
\centering
\begin{tabular}{c|llr|llr}
\toprule
\multirow{2}{*}{Requests}    &
\multicolumn{3}{c}{PROTES-1} &
\multicolumn{3}{c}{PROTES-2} \\
\cline{2-7}
&
Text & Translation & Loss &
Text & Translation & Loss \\
\midrule

50
& \foreignlanguage{russian}{ощуъиъв}
& Feelings
& -25.08
& 
& N/A
&
\\

100
& \foreignlanguage{russian}{бфзскйт}
& bfzskyt
& -20.66
& \foreignlanguage{russian}{ъщущчны}
& Synopsis
& -29.56
\\

150
& \foreignlanguage{russian}{бреощее}
& Breaking
& -27.50
& 
& N/A
&
\\

200
& \foreignlanguage{russian}{гбьъьиэ}
& gbjie
& -24.08
& \foreignlanguage{russian}{лзйшеже}
& better
& -34.01
\\

250
& \foreignlanguage{russian}{бёацсжю}
& boatsjue
& -23.15
& 
& N/A
&
\\

300
& \foreignlanguage{russian}{зсзгвлэ}
& ssgvle
& -30.42
& \foreignlanguage{russian}{едущпяз}
& Going
& -31.05
\\

350
& \foreignlanguage{russian}{ёренщял}
& fucking
& -19.84
& 
& N/A
&
\\

400
& \foreignlanguage{russian}{бфйтйвф}
& bfjtjvf
& -23.08
& \foreignlanguage{russian}{ждкнжюю}
& waiting for
& -32.49
\\

450
& \foreignlanguage{russian}{иьллтет}
& yyllt
& -18.23
& 
& N/A
&
\\

500
& \foreignlanguage{russian}{пслсждб}
& pslsjdb
& -28.03
& \foreignlanguage{russian}{лоюоыыф}
& looyouyf
& -23.54
\\

550
& \foreignlanguage{russian}{рбэхеёе}
& rbhehehehehe
& -22.68
& 
& N/A
&
\\

600
& \foreignlanguage{russian}{аэждяэй}
& aejay
& -16.74
& \foreignlanguage{russian}{псжфйбз}
& psjfybz
& -27.24
\\

650
& \foreignlanguage{russian}{быншийя}
& former
& -34.84
& 
& N/A
&
\\

700
& \foreignlanguage{russian}{сахкььй}
& Sahkyy
& -19.91
& \foreignlanguage{russian}{ёсычвжь}
& urchin
& -42.89
\\

750
& \foreignlanguage{russian}{кццьаъг}
& ktsuag
& -19.19
& 
& N/A
&
\\

800
& \foreignlanguage{russian}{клчочлй}
& klcholy
& -24.74
& \foreignlanguage{russian}{бкдммсд}
& bcdmsd
& -26.14
\\

850
& \foreignlanguage{russian}{ёбсышчн}
& Fucking
& -31.27
& 
& N/A
&
\\

900
& \foreignlanguage{russian}{йьръжиь}
& yrzhi
& -21.52
& \foreignlanguage{russian}{щуэёдьу}
& squeeze
& -32.59
\\

950
& \foreignlanguage{russian}{ёёщеяйк}
& urchin
& -30.73
& 
& N/A
&
\\

1000
& \foreignlanguage{russian}{чотёайь}
& READ MORE
& -35.34
& \foreignlanguage{russian}{счеочье}
& account
& -32.32
\\

\bottomrule
\end{tabular}
\end{table}

We consider three well-known online translators DeepL, Google, and Yandex, and search for hallucinogens following the scheme presented in the previous section.
For each translator, we limit the optimizer budget to $m = 1000$ translations and use the default values for the rest of the parameters.

Results\footnote{
    As of this writing, all of the results presented for DeepL and Yandex (and Figure~\ref{fig:demo_manual_google} for Google) can be reproduced in a modern web browser.
    The results (see Tables~\ref{tbl:results_google_level1} and~\ref{tbl:results_google_level2}) for Google translator were obtained with an older version of the browser (Chrome Canary 111.0.5555.0), which loads an older version of the translator, and are not fully reproducible in modern web browsers.
} for DeepL, Google and Yandex are presented in Tables~\ref{tbl:results_DeepL_level1},~\ref{tbl:results_google_level1} and~\ref{tbl:results_yandex_level1}, respectively.
Note that using the found seven-letter hallucinogens in Russian, we can easily manually build funny examples for each of the translators, in which the junk text at the input is translated into the correct text in English.
Please, see the related examples in Figures~\ref{fig:demo_manual_DeepL},~\ref{fig:demo_manual_google} and~\ref{fig:demo_manual_yandex}.

Then we run the optimization process for the phrases of top-$7$ hallucinogens from the first stage.
The corresponding results are presented in Tables~\ref{tbl:results_DeepL_level2},~\ref{tbl:results_google_level2} and~\ref{tbl:results_yandex_level2}.
Note that optimization based on perplexity, in this case, yields phrases that are translatable into English, but not always expressive enough (the complete list of phrases is presented in our repository).
Therefore, in the tables, we report three hand-selected quite expressive results for each of the translators.

The same procedure is conducted for the DeepL translator with the generation of longer sequences of hallucinogens.
In this case, we use the top-$33$ phrases of $7$ hallucinogens from the results of the second step, and, as before, compose their combinations of length $7$ (that is, in this case we are making a sequence of hallucinogens of length $49$). 
As a result, an interesting fact was discovered: DeepL fails to translate back into Russian the obtained meaningful English phrases.
In Figures~\ref{fig:DeepL_attack1}--\ref{fig:DeepL_attack5} we report some related examples of the adversarial attacks.

\paragraph{Parameters of the optimizer.}
In all experiments, we used the default set of parameters for PROTES (below we will call this configuration ``PROTES-1''): $K = 50$ (the number of generated samples per iteration, i.e., the batch size), $k = 5$ (the number of selected candidates per iteration), $k_{gd} = 100$ (the number of gradient ascent steps), $\lambda = 10^{-4}$ (the gradient ascent learning rate), $R = 5$ (the TT-rank of the probability tensor), and we limit the number of requests to the translator at the value  $m = 10^3$.
To evaluate the influence of the choice of parameters on the final result, we also try the following configuration (``PROTES-2''): $K = 100$, $k = 10$, $k_{gd} = 1$, $\lambda = 0.05$, $R = 5$.

To compare two sets of parameters\footnote{
    Our choice of configurations ``PROTES-1'' and ``PROTES-2'' corresponds to the parameters used in the first and second versions of the original work~\cite{batsheva2023protes}.
} we consider the DeepL online translator, and in Table~\ref{tbl:results_deepl_check} we present the best-generated hallucinogens for each requested batch (that is, for every batch of $50$ and $100$ inputs for translation requested by the optimizer ``PROTES-1'' and ``PROTES-2'', respectively).
In Figure~\ref{fig:check} we present the dependence of the found optimum (i.e., the value of the loss function) on the number of requests and related distributions for ``PROTES-1'' and ``PROTES-2''.
As can be seen from the above results, the second optimizer configuration gives better quality results, but in both cases, the successful generation of hallucinogens occurs.
Thus, our problem of generating adversarial attacks is successfully solved on the default optimizer parameters. However, as follows from the convergence curves in Figure~\ref{fig:check}, if there are more impressive budgets for requests to the translator further improvement of the results is possible.
\section{Related work}

In recent years, large language models have improved significantly in various NLP areas, especially in generative tasks.
A lot of new concepts were introduced, starting from attention mechanism~\cite{bahdanau2014neural}, transformers~\cite{vaswani2017attention} to multitask, learning from instructions~\cite{wang2022super} and human feedback~\cite{wang2021putting}.
The last becomes extremely popular in the generative context including machine translation. 
% new architectures were proposed~\cite{radford2019language,brown2020language}, and, 
Consequently, the usage of machine translation tools has become a necessary compound for understanding a foreign language. 
Unfortunately, like other neural network-based algorithms, these tools are vulnerable to adversarial examples~\cite{DBLP:journals/corr/GoodfellowSS14}. 
Starting from text classification \cite{li-etal-2020-bert-attack,DBLP:conf/acl/EbrahimiRLD18,Li2018TextBuggerGA}, vulnerability and robustness received a lot of attention in the NLP community. 
For MT systems one of the pioneering works was~\cite{ebrahimi2018adversarial}, where a character-level approach to generate adversarial examples was proposed.
Inheriting HotFlip~\cite{ebrahimi-etal-2018-hotflip} there were considered
% white- and black-box
settings, where only a few symbols in an input query are subject to change imitating typos.

While white-box optimization may yield stronger adversarial perturbations it implies access to the model's architecture and weights which is impractical in the case of online MT tools. 
In~\cite{wallace} there was considered a white-box universal approach to a targeted attack on conditional text generation. 
The authors modeled perturbation as an insertion of a trigger, a token sequence of small length, that results in a generated sequence similar to the target set of sentences. 
While during experiments certain triggers cause a model to produce sensitive racist output, they are generally meaningless and similarly to character-level attacks are easy to detect. 
Authors of~\cite{guo-etal-2021-gradient,9747475} reported high attack transferability making this approach promising for black-box setup, however,  the research is limited only to the GPT-2 model for generation task.
The above papers use greedy techniques to walk through the searching space during the optimization, on the other hand, attacks on NLP models could be found via projection onto embeddings~\cite{wallace}, and for MT task this was discovered in~\cite{Seq2Sick,Sadrizadeh2023TargetedAA,sadrizadeh2023transfool}. 
In~\cite{zhang2021crafting}, it was shown that black-box optimization may yield transferable word-level attack that fools online translation tools, e.g., Baidu and Bing. 
This work proposed to use the word saliency as the measure of uncertainty. 
Masking candidates the saliency was estimated via additional BERT model~\cite{devlin2018bert} which lead to strong readable and imperceptible adversaries, however, neither human evaluation was performed nor quantities results for online tools were given. In~\cite{wan2022paeg}, a gradient-based approach to generate phrase-level adversarial examples for neural MT systems was proposed. Similarly to~\cite{zhang2021crafting}, it is proposed to estimate the vulnerable word positions are estimated in an input phrase with the use of gradient information and replace corresponding words by the candidates computed with an auxiliary model.

We also note the recent work~\cite{guerreiro2022looking}, in which the hallucination problem of MT systems is discussed and the method for detecting and alleviating such hallucinations is presented.
Authors identified a set of hallucinations in a large number of translations by various hallucination detection methods (anomalous encoder-decoder attention, simple model uncertainty measures, etc.), and gathered for them human annotations.
This allowed them to conduct a comparative analysis of detection methods and to suggest a new approach for detection.
\section{Conclusion}

In this work, we propose a simple and effective approach to generate hallucinogens -- nonsensical gibberish  in one language that is translatable into another language by online translation tools. We evaluated our method on popular online translation systems -- Google, DeepL, and Yandex.  We found out that such systems process adversarial examples unpredictably: they not only translate nonsensical input in Russian but also can not translate seemingly meaningful English phrases.  This vulnerability may interfere with understanding a new language and worsen user’s experience while using machine translation systems, hence, additional improvements of these tools are required to establish better translation.
\section*{Acknowledgements}
    This work was supported by the Ministry of Science and Higher Education of the Russian Federation (Grant No. 075-15-2020-801).
    AC would like to thank Lev Chertkov for discovering the possibility of successful adversarial attacks on online translators using the translation result for a set of hallucinogens.
\bibliographystyle{helpers/splncs04}
\bibliography{biblio}
\end{document}